\documentclass{article}
\usepackage{comment}
\usepackage{spconf,amsmath,amsfonts,graphicx}
\usepackage{adjustbox}
\usepackage{algorithm}
\usepackage{algpseudocode}
\usepackage{enumitem}
\usepackage{adjustbox}
\usepackage{multirow}
\usepackage{cite}
\allowdisplaybreaks

\title{Deep fusion of multi-object densities using transformer }
\name{Lechi Li, Chen Dai, Yuxuan Xia, Lennart Svensson}
\address{Chalmers University of Technology, Gothenburg, Sweden}
\begin{document}
\ninept
\maketitle
\begin{abstract}
The fusion of multiple probability densities has important applications in many fields, including, for example, multi-sensor signal processing, robotics, and smart environments. In this paper, we demonstrate that deep learning based methods can be used to fuse multi-object densities. Given a scenario with several sensors with possibly different field-of-views,  tracking is performed locally in each sensor by a tracker, which produces random finite set multi-object densities. To fuse outputs from different trackers, we adapt a recently proposed transformer-based multi-object tracker, where the fusion result is a global multi-object density, describing the set of all alive objects at the current time. We compare the performance of the transformer-based fusion method with a well-performing model-based Bayesian fusion method in several simulated scenarios with different parameter settings using synthetic data. The simulation results show that the transformer-based fusion method outperforms the model-based Bayesian method in our experimental scenarios. The code is available at {https://github.com/Lechili/DeepFusion.}
\end{abstract}

\section{Introduction}\label{sec:intro}
\vspace{-2mm}
In modern surveillance systems or wireless sensor networks, multiple sensors with possibly different field-of-views (FoVs) are often utilized to provide state estimates of moving objects in the whole region of interest \cite{da2021recent}. To leverage the information available at local sensors in a distributed fashion, the multi-object densities (MODs), which capture all the information of interest of objects in the sensors' FoVs, need to be fused. Two popular approaches for fusing MODs are the geometric average (GA) fusion \cite{mahler2000optimal} (also known as logarithmic opinion pool \cite{genest1986combining}) and the arithmetic average (AA) fusion \cite{li2020arithmetic} (also known as the linear opinion pool \cite{genest1986combining}). Both fusion methods aim to find the global fused MOD that minimizes a certain loss function from the local MODs to be fused, while avoiding double counting the common information, which is essential in multi-sensor fusion. However, neither the GA nor the AA rule provides the Bayesian optimal fusion results \cite{da2021recent}. Further analysis regarding the rational and relevance of densities in multi-object/multi-sensor contexts and their guarantees of estimation efficiency may be found in \cite{braca2013asymptotic,koliander2022fusion}.

In recent years, random finite sets (RFSs) based multi-object tracking methods are gaining momentum \cite{da2021recent}, and many applications of GA and AA rules to the fusion of RFS type MODs for sensors with different FoVs have been presented \cite{gao2020fusion,gao2022fusion,yi2020distributed,uney2019fusion,yi2021heterogeneous,li2022best}. However, these suboptimal fusion methods only make use of the local multi-object posterior densities at the current time, while ignoring information regarding the objects' previous states. As discussed in \cite{chong2015track}, improved fusion results can be obtained by considering the fusion of augmented state estimates at multiple times (i.e., object trajectories). A similar idea has been applied to fusing of random finite sets of trajectories \cite{wang2022robust}, where the track association problem is addressed based on the trajectory metric \cite{garcia2020metric}. Though the method in \cite{wang2022robust} presents reasonable fusion results, it is computationally inefficient and does not make full use of all the uncertainties captured in the MOD on sets of trajectories \cite{garcia2019multiple}. Developing a model-based Bayesian fusion method that can leverage the full multi-trajectory densities well is certainly a non-trivial task. Therefore, in this paper the objective is instead to solve the MOD fusion problem in a data-driven fashion using deep learning. 

In recent years, revolutionary advances in deep learning (DL) have brought opportunities for combining DL with information fusion in multi-object tracking (MOT) \cite{blasch2021machine}. In particular, deep multi-object trackers based on transformers \cite{vaswani2017attention} have shown excellent performance in many applications, including both the model-free setting with high-dimensional data such as images and video data \cite{meinhardt2022trackformer,zeng2021motr}, and the model-based setting with low-dimensional data \cite{pinto2021next,pinto2022can}. We observe that the two problems MOT and fusion of MODs share some similarities. First, both of them can be treated as a set prediction problem where the desired output is a set of parameter representations that describes the object states of interest. Second, the input of both tasks involves time sequences of data capturing spatio-temporal information. Specifically, the input of MOT is a sequence of measurement sets, whereas the input of MOD fusion are sets of trajectories (state sequences) density parameters. In light of this, it is worth investigating how to adapt current transformer architectures for MOT to fusion of MODs. In this paper, we address the problem of fusion of MODs using deep learning, a research topic that has not been explored in current literatures. Concretely, we apply a recently developed transformer-based multi-object tracker MT3v2 \cite{pinto2022can} to the fusion of MODs. Compared to many existing architectures for deep MOT, MT3v2 is good at capturing the long-range patterns of the input sequence and allows the model to make content-aware predictions with uncertainties, making it suitable to the problem setting of fusion of MODs. The main contributions of this paper include:
\begin{itemize}[leftmargin=*]

    \item The first deep learning based solution to the fusion of MODs.
    
    \item  We adapt MT3v2 to fuse local multi-trajectory densities, computed using sensors with different FoVs, to obtain a global MOD that describes the set of current object.

    \item We compare the fusion performance of MT3v2 with a well-performing Bayesian fusion method in a model-based setting using synthetic data in a simulation study with several different parameter settings. 
    
\end{itemize}

\vspace{-2mm}
\section{Problem Formulation}\label{Problem Formulation}
\vspace{-2mm}
We consider a problem setting with $S$ sensors located in different positions with partially overlapped FoVs. These sensors jointly surveil an environment where objects may move across different sensor FoVs. The ground truth trajectories and measurements are simulated using the standard multi-object dynamic and measurement models for point objects \cite{mahler2007statistical}. For the multi-object dynamic model, each object $x_t \in \mathbb{R}^4$ at time step $t$ survives with probability $p^s$ and moves independently according to a constant velocity motion model. Newborn objects appear following a Poisson point process (PPP) with Poisson birth rate $\lambda^b$, and the birth density is a single Gaussian with a large covariance covering the whole area of interest. As for the multi-object measurement model, each object $x_t$ is detected with probability $p^d$, and if detected, it generates a single measurement $z_t \in \mathbb{R}^2$ according to a linear Gaussian measurement model. For the $s$-th sensor with $s \in \{1,...,S\}$, the set of measurements $\mathbf{z}^s_t$ at time step $t$ is the union of measurements generated by objects inside in its FoV and PPP clutter measurements. Further, we denote the sequence of measurement sets up to and including time step $t$ collected by the $s$-th sensor as $\mathbf{z}_{1:t}^s$ and the set of trajectories at time step $t$ as $\mathbf{X}_t$. 

For the standard multi-object dynamic and measurement models, the closed form solution for estimating sets of trajectories is given by the trajectory Poisson multi-Bernoulli (TPMB) mixture filter \cite{granstrom2018poisson}. The TPMB filter \cite{garcia2020trajectory} is an efficient approximation of the TPMB mixture filter, where the multi-trajectory posterior density is represented in a simpler form with much fewer parameters. At the $s$-th sensor, a TPMB filter is used to process the measurements and recursively compute the approximate multi-trajectory posterior density $f^s(\mathbf{X}_t | \mathbf{z}_{1:t}^s)$ in the form of a PMB. In a TPMB density, the set of undetected trajectories is represented by a PPP, and the set of detected trajectories is represented by a multi-Bernoulli (MB). Each Bernoulli component in the trajectory MB describes a potential detected trajectory and is characterized by an existence probability and a single trajectory density. 

To recursively estimate the global multi-object density of $\mathbf{x}_t$, MT3v2 can be applied in a sliding window fashion, as described in \cite{pinto2022can}. In this paper, we consider a simpler setting, where all the local trajectory MB densities at the final time step $T$ are fused to obtain the global multi-object density $f(\mathbf{x}_T | \mathbf{z}_{1:T}^1,...,\mathbf{z}_{1:T}^{S})$ representing the set of detected objects $\mathbf{x}_T$ at time step $T$. The desired global multi-object density is also an MB, and each Bernoulli component is characterized by an existence probability and a single object density. 

\vspace{-2mm}
\section{Method}\label{Method}
\vspace{-2mm}
In this section, we first give an overview of the structure of MT3v2 \cite{pinto2022can}. Then, we describe how to prepare the input data to MT3v2 using trajectory MB densities obtained at local sensors, including input embeddings and positional encoding. This is the main difference compared to the implementation in \cite{pinto2022can} where the input data is a sequence of measurement sets. Finally, we briefly describe the methods used to evaluate the fusion performance.
\subsection{MT3v2}
\begin{figure}[t!]
\centerline{\includegraphics[width=0.7\linewidth]{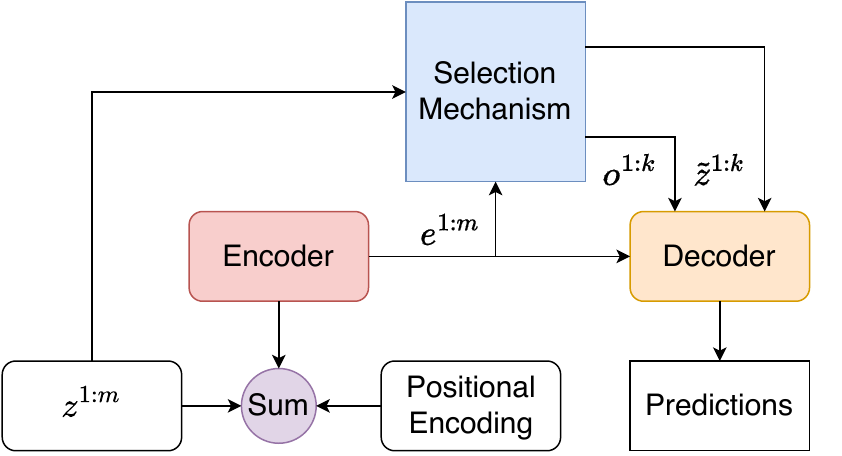}}
\caption{\footnotesize The overall structure of MT3v2.}
\vspace{-5mm}
\label{fig: mott}
\end{figure}
An overview of the structure of the MT3v2 \cite{pinto2022can}, is shown in Fig. \ref{fig: mott}. Based on the transformer architecture \cite{vaswani2017attention}, MT3v2 contains an encoder and a decoder. The encoder first maps the measurements $z^{1:m}$ (obtained by concatenating all the measurements in $\mathbf{z}_{1:T}$) to embeddings $e^{1:m}$. The positional encoding \cite{vaswani2017attention} outputs a sequence of vectors, which typically have the same shape as the input, and hence they can be added together, to make the model aware of the temporal information of the measurement sequence. After the encoder, we feed embeddings $e^{1:m}$, together with initial predictions $\tilde{z}^{1:k}$ and object queries $o^{1:k}$ produced by a selection mechanism \cite{pinto2021next,pinto2022can}, to the decoder to obtain the predictions of the object states. The decoder of the MT3v2 is similar to the one used in the DEtection TRansformer (DETR) \cite{carion2020end}, and it uses a process called iterative refinement \cite{pinto2022can} to adjusts its predictions at each decoder layer. Further, MT3v2 is trained by optimizing the negative log likelihood (NLL) of the MB density evaluated at the ground truth \cite{pinto2022can}.

\subsection{Data preparation}\label{sec:datapar}
The tracker located at each sensor node outputs a distribution for each trajectory, with parameters describing local trajectory MB densities, which needs to be transformed into a sequence of vectors in order to be used by the transformer model. To do so, we produce one vector for each sensor and object at a specific time steps, using the information contained in these parameters. In particular, for the linear Gaussian TPMB implementation \cite{garcia2020trajectory} at the $s$-th sensor, $s \in \{1,\dots,S\}$, its output at time step $T$ is a trajectory MB density with $n^s$ Bernoulli components, where the $i$-th Bernoulli, $i \in \{1,\dots,n^s\}$, is described by the following parameters:
\begin{itemize}
\item Existence probability $r^{s,i}\in(0,1]$.
\item Trajectory start time $t^{s,i}$.
\item Trajectory maximum length $\ell^{s,i}$.
\item Trajectory length probability vector of length $\ell^{s,i}$ where the $j$-th element $w^{s,i}_j \in \mathbb{R}^{\ell^{s,i}}, j\in\{1,\dots,\ell^{s,j}\}$, represents the probability that the trajectory has length $j$. 
\item Mean of state sequence $x_{1:\ell^{s,i}}^{s,i} \in \mathbb{R}^{4\ell^{s,i}}$.
\item Covariance of state sequence $P_{1:\ell^{s,i}}^{s,i} \in \mathbb{R}^{4\ell^{s,i} \times 4\ell^{s,i}}$.
\end{itemize}
Here, superscripts $s,i$ are used to indicate that the quantity of interest is related to the $i$-th Bernoulli component in the MOD reported by the $s$-th sensor. In addition, subscript $j$ is used to indicate that the quantity of interest is related to the $j$-th element (object state) in the state sequence density of a Bernoulli component.

Directly using all these parameters as input data can be space and time-consuming, and reducing the number of parameters is therefore needed. To do so, we first prune Bernoulli components with existence probability less than a certain threshold $r^{s,i} < p_{\text{Ber}}$, and we denote the number of Bernoulli components after pruning as $\hat{n}^s$. Second, for a sequence of object states $x_{1:\ell^{s,i}}^{s,i}$, we ignore the correlation between object states at different time steps, and extract covariance matrices $\{C_{j}^{s,i}\}_{j=1}^{\ell^{s,i}}$ from $P_{1:\ell^{s,i}}^{s,i}$, where $C_{j}^{s,i} \in \mathbb{R}^{4\times 4}$ corresponds to the sub-matrix of $P_{1:\ell^{s,i}}^{s,i}$ with row and column number from $4j-3$ to $4j$, $ j\in\{1,\dots,\ell^{s,j}\}$. Since $C$ is symmetric and positive finite, we select the entries along and above the main diagonal of a matrix $C\in\mathbb{R}^{4\times4}$,  followed by arranging them to a vector $c\in\mathbb{R}^{10}$ in order from left to right and from top to bottom. Further, we use $\hat{w}^{s,i} = r^{s,i}w^{s,i}$ to represent the marginal object existence probability $\hat{w}^{s,i}_j$ at time step $t^{s,i}+j-1$. Finally, for the $i$-th trajectory Bernoulli component obtained from the $s$-th sensor, the parameters describing the object state at time step $t^{s,i}+j-1, j \in \{1,\dots,\ell^{s,i}\}, $ can be represented using a column vector $u = [ x^{s,i}_j; c^{s,i}_j;\hat{w}^{s,i}_j] \in\mathbb{R}^{15}$. 

To account for the case when sensors are mobile, the object state vector $u$ can be augmented to include the sensor position information. Let $a_j^s \in R^2$ and $b^s \in R$ (time-invariant) be the position and orientation of the $s$-th sensor corresponding to object state $x_j^{s,i}$. The parameter representation of the object state at time step $t^{s,i} + j - 1$ now becomes $u = [ x^{s,i}_j; c^{s,i}_j; a^{s}_j ; b^{s} ;\hat{w}^{s,i}_j]\in\mathbb{R}^{18}$. Using this representation, we can transform the parameters describing the S trajectory multi-Bernoulli densities into a sequence of vectors $\{u_k\}_{k=1}^l$, and the length of the sequence is given by $l = \sum_{s=1}^{S} \sum_{i=1}^{\hat{n}^s} \ell^{s,i}$.

\subsection{Input embeddings and positional encoding}
We proceed to describe the input embeddings and how to encode the time step $t \in \{t^{s,i},\cdots,t^{s,i}+\ell^{s,i}-1\}$, the trajectory index $i \in \{1,\cdots,\ell^{s,i}\}$, and the sensor index $s \in \{1,\cdots,S\}$ to the input embeddings. We first normalize vectors in $\{u_k\}_{k=1}^l$ using FoVs information, followed by augmenting their dimensionality to $d^\prime$ through a linear transformation \cite{pinto2021uncertainty}, where $d^\prime = 256$ is a hyperparameter. The resulting sequence is denoted by $\{\hat{u}_k\}_{k=1}^l$. In addition, we define three positional encoders $\phi^{\text{Time}}$, $\phi^{\text{Trajectory}}$ and $\phi^{\text{Sensor}}$, which all map an index to a $d^\prime$-dimensional vector. The positional encoders are lookup tables that contains learnable embeddings of fixed length and size. After positional encoding, each vector of $\{\hat{u}_k\}_{k=1}^l$ becomes 
\begin{equation}
    u_{k}^{\prime} = \hat{u}_{k} + \phi^{\text{Time}}(t) + \phi^{\text{Trajectory}}(j) + \phi^{\text{Sensor}}(n) \in \mathbb{R}^{d^\prime}.
    \end{equation}
The sequence of preprocessed input vectors is then fed to the MT3v2 encoder for further downstream operations.

\subsection{Performance evaluation and losses}
The output of MT3v2 is a multi-Bernoulli density at time step $T$, where each Bernoulli component captures uncertainties of a potential object state, parameterized by an existence probability, an object state mean vector and a covariance matrix, denoted as $\beta^{1:k} = (r^{1:k}, \mu_T^{1:k}, \Sigma_T^{1:k})$. The fusion performance of MT3v2 is compared with the Bayesian fusion method in \cite{frohle2020decentralized}, which takes the local PMB multi-object densities at time step $T$ as input and outputs the global PMB density describing the set of object states at time step $T$. The local PMB multi-object densities at time step $T$ can be obtained by marginalizing the local PMB multi-trajectory densities \cite{granstrom2018poisson}.

For evaluating the MOD fusion performance, we use the generalized optimal sub-pattern assignment (GOSPA) metric \cite{rahmathullah2017generalized}  as well as the NLL \cite{pinto2021uncertainty}. Both performance measures can be decomposed into three parts: the localization error, the misdetection error, and the false detection error, allowing for a systematic analysis of fusion errors. In addition, NLL is an uncertainty-aware performance measure, which, compared to GOSPA, is able to take into account all the uncertainties captured in the MOD. The loss function for training MT3v2 is also based on approximation of the NLL, which is defined \cite[Eq.(24)]{pinto2022can}.

\vspace{-2mm}
\section{Simulation Results}\label{sec:Simulation Results}
\vspace{-2mm}
In this section, we first introduce the simulation scenarios, followed by the implementation details for the data generation, model configuration and evaluation. Then, we show the results and present the analysis.
\subsection{Experiential setup and implementation detail}

\begin{figure}[t!]
\centering
\includegraphics[width=\linewidth]{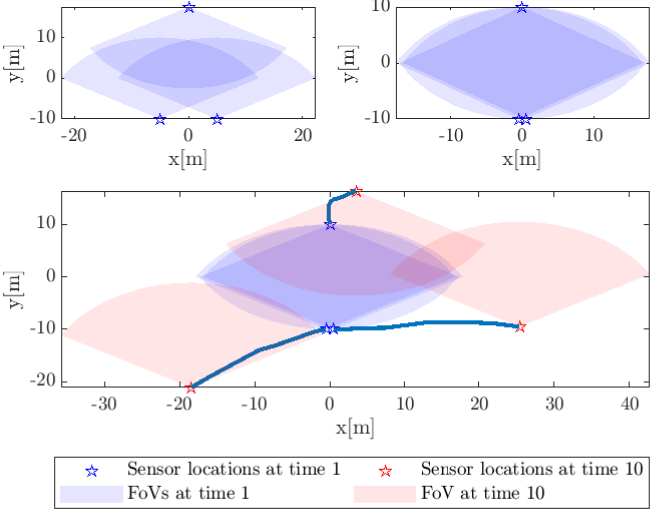}
\caption{\footnotesize An illustration of scenario 1 (top left), scenario 2 (top right), and a sample plot of scenario 3 (bottom,  sensors’ trajectories are shown in solid lines). }
\vspace{-5mm}
\label{fig: scenarios}
\end{figure}

\noindent\textbf{Scenario Setup}: The simulations are conducted in three different scenarios, shown in Fig. \ref{fig: scenarios}. In each scenario, there are three sensors, each of which has a fan-shaped FoV with a bearing size $2/(3\pi)$ and a radius 20 m. The first two scenarios have fixed the sensors, while sensors in scenario 3 are mobile, with initial sensors' locations the same as those in scenario 2. For each scenario, we define two tasks, where task 1 is a baseline and task 2 has a higher measurement noise than task 1. In addition, objects move according to a constant velocity model, and are measured by a linear Gaussian model, as described in Section \ref{Problem Formulation}. The multi-object model parameters have been summarized in Table \ref{tab:MF}. In scenario 3, the sensors also move according to a constant velocity model, but with $\sigma_q = 10$.
\begin{table}[t!]
\centering
\caption{\footnotesize Multi-object model parameters for tasks in each scenario.}
\label{tab:MF}
\begin{adjustbox}{width=0.73\linewidth}
\begin{tabular}{l|l|c|c}
\hline
& Parameter name        & Task1 & Task2 \\ \hline
\multirow{6}{*}{Parameters} & Process noise  $\sigma_q^2$       & 0.5   & 0.5   \\
& Measurement noise $\sigma_z^2$       & 0.01  & 0.1  \\
& Scan time $\Delta_t$            & 0.1   & 0.1   \\
& Birth rate $\lambda_b$            & 0.1   & 0.1   \\
& Clutter rate $\lambda_c$               & 5     & 5     \\
& Survival probability $p^s$  & 0.9   & 0.9   \\
& Detection probability  $p^d$ & 0.95  & 0.95  \\ \hline
\end{tabular}
\end{adjustbox}
\vspace{-5mm}
\end{table}

\noindent\textbf{Implementation details}:  In this work, we use synthetic data for training, validation and testing. Concretely, object states and measurements are generated according to the standard multi-object dynamic and measurement models with parameters in Table \ref{tab:MF}. Measurements are processed by local TPMB filters to produce trajectory multi-Bernoulli densities, and their parameters are transformed to a sequence of vectors as described in Section \ref{sec:datapar}. For the TPMB filter located at each sensor node, the gating size is 20. The maximum number of hypothesis is 100. In addition, we prune MBs with weight smaller than
$10^{-3}$, Bernoulli components with probability of existence
smaller than $p_{\text{th}}=10^{-3}$ and Gaussian components in the Poisson intensity for undetected objects with
weight smaller than $10^{-5}$. The existence estimation threshold controls the number of output trajectories estimates, which is set to be 0.5. Further, we collect 100k groups of data for training and 25000 for validation for each task, where each group contains 32 batches of sequence of vectors (sets of trajectories). Furthermore, MT3v2 is trained for 4 epochs (400k gradient steps) using an ADAM optimizer with initial learning rate 0.00005, and the learning rate will decrease by 1/4 if the total training losses do not decrease for 50000 gradient steps.

\begin{table*}[t!]
\caption{\footnotesize GOSPA and NLL errors along with their decomposition for scenario 1, 2 and 3.}
\label{tab:S1S2}
\begin{adjustbox}{width=\textwidth}
\begin{tabular}{c|cccc|cccc|cccc}
\hline
Scenarios   & \multicolumn{4}{c|}{1}                                 & \multicolumn{4}{c|}{2}                                 & \multicolumn{4}{c}{3}                                 \\ \hline
Tasks       & \multicolumn{2}{c}{Task1} & \multicolumn{2}{c|}{Task2} & \multicolumn{2}{c}{Task1} & \multicolumn{2}{c|}{Task2} & \multicolumn{2}{c}{Task1} & \multicolumn{2}{c}{Task2} \\ \hline
Methods     & Bayesian & MT3v2          & Bayesian  & MT3v2          & Bayesian & MT3v2          & Bayesian  & MT3v2          & Bayesian & MT3v2          & Bayesian & MT3v2          \\ \hline
GOSPA-total & 1.618    & \textbf{1.373}          & 2.110     & \textbf{2.019} & 1.440    & \textbf{1.413} & 2.007     & \textbf{1.933} & 1.294    & \textbf{0.885}          & 1.578    & \textbf{1.312} \\
GOSPA-loc   & 0.596    & 0.959          & 0.821     & 1.492          & 0.693    & 1.040          & 1.028     & 1.471          & 0.256    & 0.597          & 0.214    & 0.947          \\
GOSPA-miss  & 0.058    & 0.051          & 0.058     & 0.149          & 0.076    & 0.031          & 0.098     & 0.154          & 0.003    & 0.025          & 0.010    & 0.090          \\
GOSPA-false & 0.964    & 0.363          & 1.231     & 0.378          & 0.671    & 0.342          & 0.881     & 0.308          & 1.035    & 0.263          & 1.353    & 0.275          \\
NLL-total   & 12.207   & \textbf{1.195} & 11.277    & \textbf{6.134}          & 2.695    & \textbf{1.833}          & 7.257     & \textbf{4.552}          & 13.105   & \textbf{0.429} & 12.591   & \textbf{3.075}          \\
NLL-loc     & 11.766   & 0.572          & 10.820    & 5.475          & 2.103    & 1.169          & 6.232     & 3.898          & 12.656   & 0.089          & 12.445   & 2.618          \\
NLL-miss    & 0.320    & 0.304          & 0.102     & 0.289          & 0.253    & 0.411          & 0.083     & 0.365          & 0.428    & 0.137          & 0.045    & 0.198          \\
NLL-false   & 0.121    & 0.319          & 0.355     & 0.370          & 0.340    & 0.253          & 0.942     & 0.289          & 0.021    & 0.202          & 0.102    & 0.266          \\ \hline
\end{tabular}
\end{adjustbox}
\vspace{-5mm}
\end{table*}


\noindent\textbf{Evaluation}: In evaluation, we pass the test data to the trained model and Bayesian fusion method, and extract those Bernoulli components from the output of the MT3v2 and Bayesian method, with corresponding existence probabilities of Bernoulli components larger than 0.75 and 0.5, respectively. This is repeated for 1000 Monte Carlo simulation runs. For the GOSPA metric, we choose the parameters $c=2$, $p=1$ and $\alpha=2$ for all tasks and scenarios. For the NLL metric, the PPP is added and tuned in the same way as in \cite[Section V]{pinto2022can}.

\begin{figure}[t!]
\centering
\includegraphics[width=0.7\linewidth]{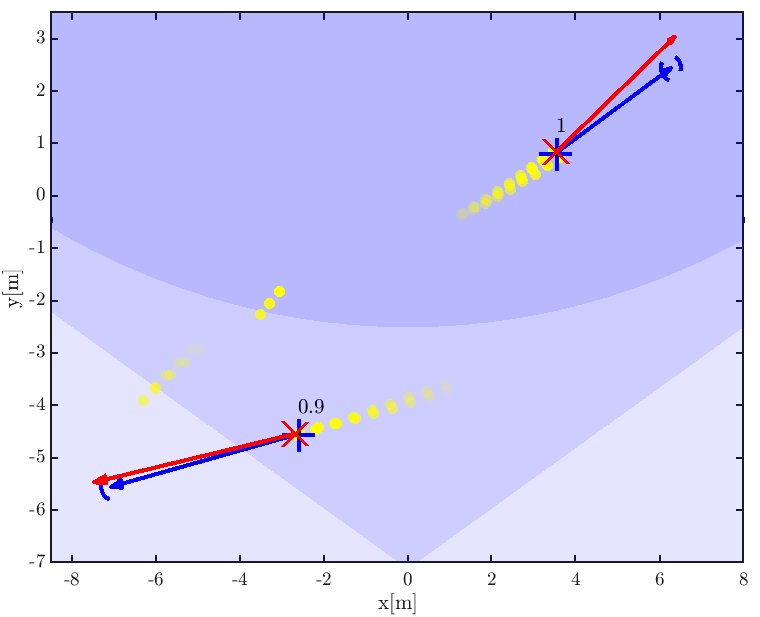}\\
\includegraphics[width=0.7\linewidth]{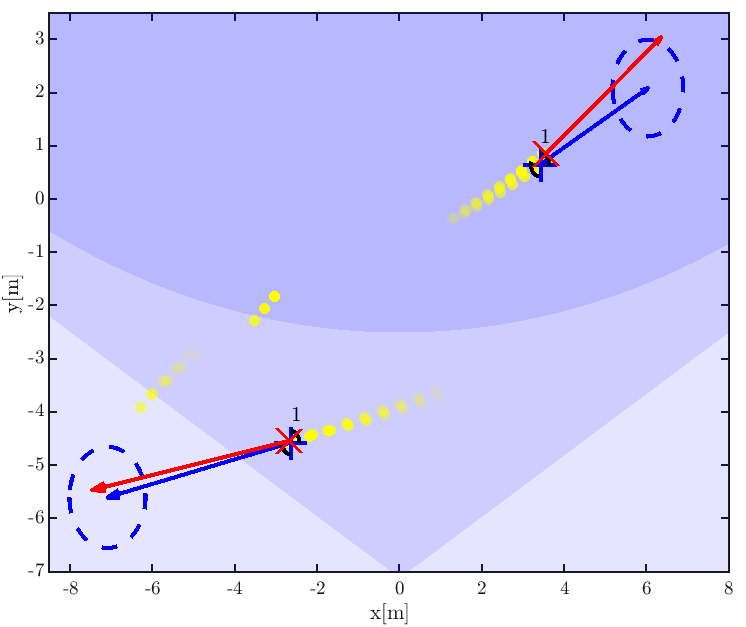}
\caption{\footnotesize An illustration  of sample plot of Bayesian method (Upper) and MT3v2 (Bottom). Yellow filled circles indicates estimated positions obtained from local filters (Circles that are more opaque represent object states closer to time step $T$). The ground-truth positions/velocities at the current time are shown in red cross/arrows,
respectively, while predicted positions/velocities are shown in blue plus signs/arrows. The numbers in each figure indicates the existence probabilities of Bernoulli components. The blue/black dished ellipses represent the 3-$\sigma$ level of predicted velocities/positions. }
\label{fig: sample_plots}
\end{figure}

\begin{figure}[t!]
\centering
\includegraphics[width=0.49\linewidth]{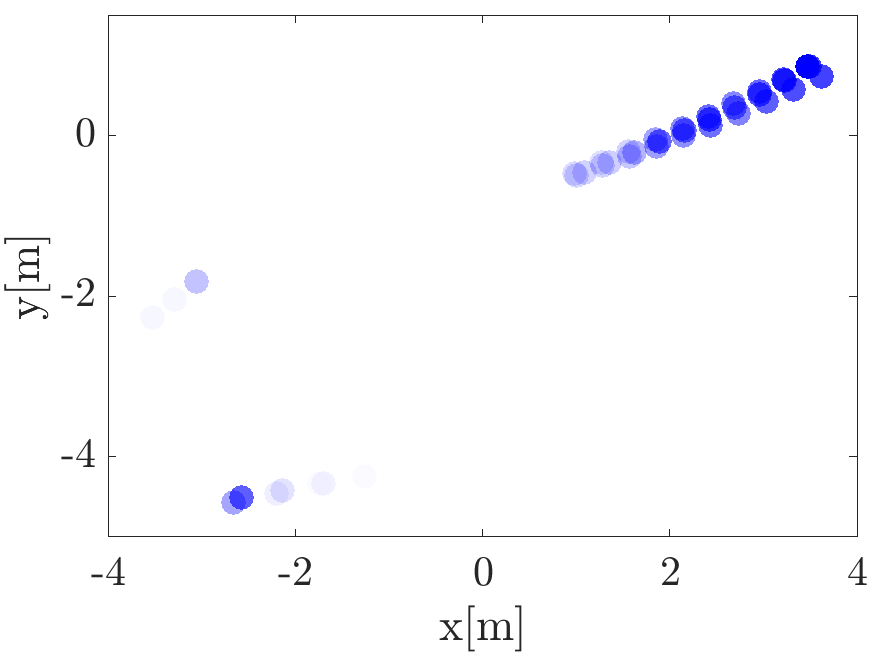}
\includegraphics[width=0.49\linewidth]{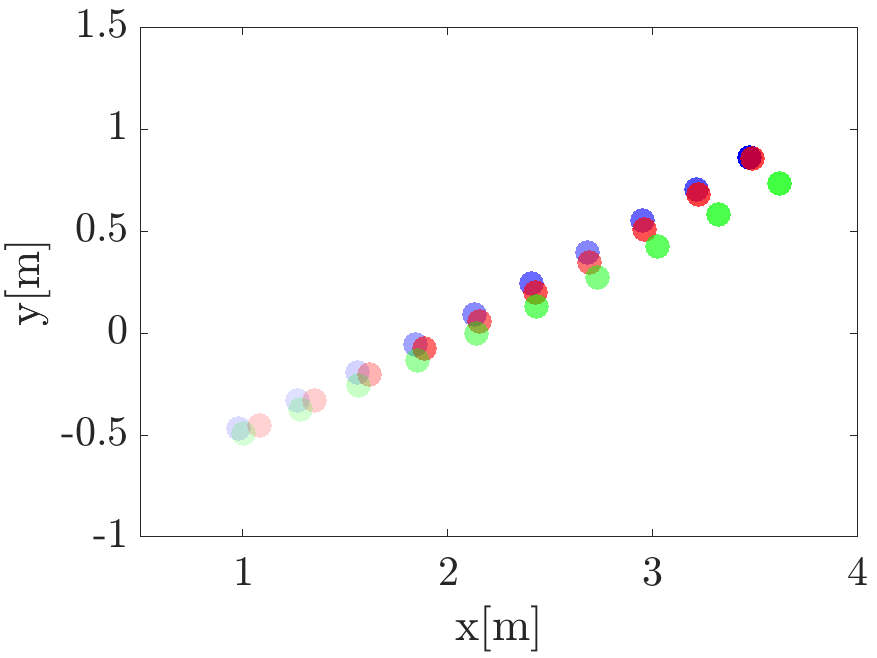}
\caption{\footnotesize In the left figure, the attention maps (corresponding to the predictions of MT3v2 in the right plot of Fig. \ref{fig: sample_plots}) are plotted with their trajectory estimates in blue-filled circles (the more opaque a circle is, the higher the attention weight is). Also, in the right figure, we plot their corresponding trajectories estimates from different sensors, indicated using different colors.}
\label{fig: attention_maps}
\vspace{-5mm}
\end{figure}
\subsection{Results analysis}

GOSPA and NLL errors along with their decomposition for scenario 1,2 and 3 are shown in Table \ref{tab:S1S2}, respectively. Judging from the average GOSPA and NLL errors, MT3v2 outperforms the Bayesian method in all the tasks and scenarios. Inspecting the decomposition of GOSPA errors, the Bayesian method has the lowest GOSPA localization errors in all tasks and scenarios, while  MT3v2 yields lower false errors. For NLL errors, MT3v2 gives lower localization errors, which means that MT3v2 provides more certain location estimates. Further, the results on task 2 show that the noise level in measurements can deteriorate the fusion performance of both methods. When the sensors are fixed (scenario 1 and scenario 2), both methods achieve a better performance as the over-lapped FoVs become larger; however, this is not observed when sensors are mobile. In scenario 3, the over-lapped FoVs often become smaller as time step increases, and the ground-truth objects at the last time step (if they exist) may only present in separate sensor's FoV. When that happens, the MOD fusion problem becomes trivial to solve.

A sample plot in scenario 1 for MT3v2 and the Bayesian method is shown in Fig. \ref{fig: sample_plots}. To gain some insights, the corresponding attention maps of MT3v2   are shown in Fig. \ref{fig: attention_maps}. From the figure, we can see that when the model makes a final prediction for a specific object, it mainly focuses on the trajectory estimates originating from that object, where local tracks are fused into a global track, while other trajectory estimates have relatively small attention weights. Further, estimates from different sensors are treated with approximately equal weights, and those from recent time steps receive greater attention than the rest. 
\vspace{-2mm}
\section{CONCLUSIONS}
\vspace{-2mm}
In this paper, we have applied MT3v2 to solve the problem of the fusion of MODs. The results on synthetic data show that the MT3v2 outperforms a well-performing model-based Bayesian fusion method in terms of both average GOSPA and NLL errors in several different simulation settings. Importantly, this work displays the potential of using deep learning based methods for the fusion of MODs.

\vfill\pagebreak

\bibliographystyle{IEEEbib}
\bibliography{mybib.bib}

\end{document}